\documentclass{article}
\usepackage{spconf,amsmath,amssymb,graphicx,color}
\usepackage{booktabs}
\usepackage{hyperref}
\usepackage{multirow}
\usepackage{lipsum}
\usepackage{float}

\title{Learned in Speech Recognition: Contextual Acoustic Word Embeddings}
\name{Shruti Palaskar$^*$, Vikas Raunak\sthanks{Equal contribution} and Florian Metze}
\address{Carnegie Mellon University, Pittsburgh, PA, U.S.A. ~\\ \{spalaska $|$ vraunak $|$ fmetze \}@cs.cmu.edu}

\begin{document}
\ninept
\maketitle
\begin{abstract}
End-to-end acoustic-to-word speech recognition models have recently gained popularity because they are easy to train, scale well to large amounts of training data, and do not require a lexicon. In addition, word models may also be easier to integrate with downstream tasks such as spoken language understanding, because inference (search) is much simplified compared to phoneme, character or any other sort of sub-word units. In this paper, we describe methods to construct contextual acoustic word embeddings directly from a supervised sequence-to-sequence acoustic-to-word speech recognition model using the learned attention distribution. On a suite of 16 standard sentence evaluation tasks, our embeddings show competitive performance against a word2vec model trained on the speech transcriptions. In addition, we evaluate these embeddings on a spoken language understanding task, and observe that our embeddings match the performance of text-based embeddings in a pipeline of first performing speech recognition and then constructing word embeddings from transcriptions.
\end{abstract}
%
%%%%%%%%%%%%%%%%%%%%%%%%%%%%%%%%%%%%%%%%
\section{Introduction}
\label{sec:intro}
The task of learning fixed-size representations for variable length data like words or sentences, either text or speech-based, is an interesting problem and a focus of much current research. In the natural language processing community, methods like word2vec~\cite{word2vec}, GLoVE~\cite{glove}, CoVe~\cite{cove} and ELMo~\cite{elmo} have become increasingly popular, due to their utility in several natural language processing tasks. Similar research has progressed in the speech recognition community, where however the input is a sequence of short-term audio features, rather than words or characters. Therefore,  the variability in speakers, acoustics or microphones for different occurrences of the same word or sentence adds to the challenge. 

Prior work towards the problem of learning word representations from variable length acoustic frames involved either providing word boundaries to align speech and text~\cite{speech2vec}, or chunking (``chopping'' or ``padding'') input speech into fixed-length segments that usually span only one word~\cite{kamper_acoustic_word_embeddings, bengio_speech_embeddings, harwath_glass_2015, he2016multi}. Since these techniques learn acoustic word embeddings from audio fragment and word pairs obtained via a given segmentation of the audio data, they ignore the specific audio context associated with a particular word. So the resulting word embeddings do not capture the contextual dependencies in speech. In contrast, our work constructs individual acoustic word embeddings grounded in utterance-level acoustics. 

In this paper, we present different methods of obtaining acoustic word embeddings from an attention-based sequence-to-sequence model \cite{seq2seq,LAS,chorowski2015attention} trained for direct Acoustic-to-Word (A2W) speech recognition \cite{palaskar2018acoustic}. Using this model, we jointly learn to automatically segment and classify input speech into individual words, hence getting rid of the problem of chunking or requiring pre-defined word boundaries. As our A2W model is trained at the utterance level, we show that we can not only learn acoustic word embeddings, but also learn them in the proper context of their containing sentence. We also evaluate our contextual acoustic word embeddings on a spoken language understanding task, demonstrating that they can be useful in  non-transcription downstream tasks.

Our main contributions in this paper are the following: ~\\1. We demonstrate the usability of attention not only for aligning words to acoustic frames without any forced alignment but also for constructing Contextual Acoustic Word Embeddings (CAWE). ~\\2. We demonstrate that our methods to construct word representations (CAWE) directly from a speech recognition model are highly competitive with the text-based word2vec embeddings \cite{word2vec}, as evaluated on 16 standard sentence evaluation benchmarks. ~\\3. We demonstrate the utility of CAWE on a speech-based downstream task of Spoken Language Understanding showing that pretrained speech models could be used for transfer learning similar to VGG in vision \cite{vgg} or CoVe in natural language understanding \cite{cove}.

%In the next section, we survey the related works in A2W modeling, as well as learning vector representations for speech. Our A2W speech recognition model is explained in Section \ref{sec:model_a2w}. Our methods to construct Contextual Acoustic Word Embeddings (CAWE) from the trained A2W speech recognition model are explained in Section \ref{sec:cawe}. Experimental results and conclusions follow thereafter.

%%%%%%%%%%%%%%%%%%%%%%%%%%%%%%%%%%%%%%%%
\section{Related Work}
\label{sec:related_work}
A2W modeling has been largely pursued using Connectionist Temporal Classification (CTC) models \cite{ctc} and Sequence-to-Sequence (S2S) models \cite{seq2seq}. Prior work shows the need for large amounts of training data for these models (thousands of hours of speech) with large word vocabularies of frequently occurring words \cite{hasim-2016-a2w, microsoft-2018-a2w,google_asr_sota,asr_bpe_google,zeyer2018improved}. Progress in the field showed the possibility of training these models with smaller amount of data (300 hours Switchboard corpus \cite{switchboard_corpus}) but restricting the vocabulary to words occurring atleast 5 or 10 times \cite{ibm_building_competitve,ueno2018acoustic}. The solutions to generate out-of-vocabulary words have revolved around backing off to smaller units like characters or sub-words \cite{ibm_building_competitve, microsoft-2018-a2w, google_asr_sota,ueno2018acoustic,zeyer2018improved}. While this solves the problem of rare word generation, the models are no longer pure-word models.

\cite{renals_seq2seq_swbd} present one of the first S2S models for pure-word large vocabulary A2W recognition with the 300 hour Switchboard corpus with a vocabulary of about 30,000 words. \cite{chen2018modular,palaskar2018acoustic} build upon their work and improve the training of these models for the large vocabulary task. \cite{palaskar2018acoustic} is one of our previous works where we show that the direct A2W model is also able to automatically learn word boundaries without any supervision and is the current best pure-word S2S model. We use the same model in this work and expand it towards learning acoustic embeddings. 

\cite{speech2vec,kamper_acoustic_word_embeddings,harwath_glass_2015,bengio_speech_embeddings,yuan2018learning,he2016multi} all explore ways to learn acoustic word embeddings. All above methods except \cite{bengio_speech_embeddings} use unsupervised learning based methods to obtain these embeddings where they do not use the transcripts or do not perform speech recognition. \cite{bengio_speech_embeddings} use a supervised Convolutional Neural Network based speech recognition model but with short speech frames as input that usually correspond to a single word. This is the common practice in most prior work that simplifies training but prevents the models to scale to learn contextual word embeddings grounded in utterance level acoustics. \cite{speech2vec} propose an unsupervised method to learn speech embeddings using a fixed context of words in the past and future. The drawbacks of their method are the fixed context and need for forced alignment between speech and words for training. 

Learning text-based word embeddings is also a rich area of research with well established techniques such as \cite{word2vec,glove}. Research has further progressed into learning contextualized word embeddings \cite{cove, elmo} that are useful in many text-based downstream tasks \cite{senteval}. \cite{cove} learns contextual word embeddings from a fully trained machine translation model and depict re-use of their encoder in other downstream tasks. Our work ties A2W speech recognition model with learning contextual word embeddings from speech.

%%%%%%%%%%%%%%%%%%%%%%%%%%%%%%%%%%%%%%%%
\section{Acoustic-to-Word Recognition}
\label{sec:model_a2w}
% \subsection{Sequence-to-Sequence Model}
Our S2S model is similar in structure to the Listen, Attend and Spell model \cite{LAS} which consists of 3 components: the encoder network, a decoder network and an attention model. The encoder maps the input acoustic features vectors $\textbf{a} = (\textbf{a}_1,\textbf{a}_2,...,\textbf{a}_T)$ where $\textbf{a}_i \in \mathcal{R}^d$, into a sequence of higher-level features $\textbf{h} = (\textbf{h}_1,\textbf{h}_2,...,\textbf{h}_{T'})$. The encoder is a pyramidal (sub-sampling) multi-layer bi-directional Long Short Term Memory (BLSTM) network. The decoder network is also an LSTM network that learns to model the output distribution over the next target conditioned on sequence of previous predictions i.e. $P(\textbf{y}_l | y_{l-1}^*,y_{l-2}^*,...,y_0^*,\textbf{x})$ where $\textbf{y}* = (y_0^*,y_1^*,...,y_{L+1}^*)$ is the ground-truth label sequence. In this work, $y_i^* \in \mathcal{U}$ is from a word vocabulary. This decoder generates targets $\textbf{y}$ from $\textbf{h}$ using an attention mechanism. 

%The attention model learns an alignment weight vector between the encoding $\textbf{h}$ and the current output of decoder $\textbf{y}_l$. At each time step, the attention module computes a context vector that is fed into the decoder together with the previous ground-truth label $y_{l-1}^*$.

We use the location-aware attention mechanism \cite{chorowski2015attention} that enforces monotonicity in the alignments by applying a convolution across time to the attention of previous time step. This convolved attention feature is used for calculating the attention for the current time step which leads to a peaky distribution \cite{chorowski2015attention,palaskar2018acoustic}. Our model follows the same experimental setup and model hyper-parameters as the word-based models described in our previous work \cite{palaskar2018acoustic} with the difference of learning 300 dimensional acoustic feature vectors instead of 320 dimensional.

\begin{table*}[h]
\begin{minipage}{\linewidth}
  \caption{Comparing three methods to obtain acoustic word embeddings from an A2W model: unweighted average (U-AVG), weighted average (CAWE-W) and maximum attention (CAWE-M).}
  \label{tab:senteval_baselines}
  \centering
  \begin{tabular}{ccccccc} %{ p{6cm}  p{1cm} p{1cm}  p{1cm} p{1cm}  }
    \toprule
                & \multicolumn{3}{c} {Switchboard}      & \multicolumn{3}{c}{How2}   \\
    \cmidrule(r){2-4}\cmidrule(r){5-7}
    Dataset   & U-AVG          & CAWE-W      & CAWE-M         & U-AVG      & CAWE-W  & CAWE-M  \\
    \midrule 
    STS 2012  & 0.3230         & 0.3281     & \textbf{0.3561}        & 0.3255     & 0.3271 & \textbf{0.3648}              \\
    STS 2013  & 0.1252         & 0.1344     & \textbf{0.1969}        & 0.2070     & 0.2071 & \textbf{0.2716}              \\
    STS 2014  & 0.3358         & 0.3389     & \textbf{0.3888}        & 0.3375     & 0.3426 & \textbf{0.3940}              \\
    STS 2015  & 0.3854         & 0.3881     & \textbf{0.4275}        & 0.3852     & 0.3843 & \textbf{0.4173}        \\
    STS 2016  & 0.2998         & 0.2974     & \textbf{0.3833}        & 0.3248     & \textbf{0.3271} & 0.3159        \\
    STS B     & 0.3667        & 0.3510    & \textbf{0.4010}        & 0.3343     & 0.3440 & \textbf{0.4000}           \\
    SICK-R    & 0.5640        & 0.5800    & \textbf{0.6006}        & 0.5800     & 0.6060 & \textbf{0.6440}        \\
    \midrule
    MR        & 63.86         & 63.75     & \textbf{64.69}         & 63.46     & 63.19 & \textbf{63.64}    \\
    MRPC      & \textbf{70.67}         & 69.45     & 69.80         & 68.29     & 67.83 & \textbf{70.61}         \\
    CR        & 71.42         & 72.13     & \textbf{72.93}         & \textbf{74.12}     & 73.99 & 73.03         \\
    SUBJ      & \textbf{82.45}         & 82.22     & 81.19         & 81.48     & \textbf{81.88} & 81.01         \\
    MPQA      & \textbf{73.76}         & 73.28     & 73.75         & \textbf{74.21}     & 74.18 & 73.53         \\
    SST       & 66.45         & \textbf{66.61}     & 65.02         & 63.43     & 63.43 & \textbf{65.13}         \\
    SST-FG    & 32.81         & 32.04     & \textbf{33.53}         & 31.95     & \textbf{32.35} & 32.03         \\
    TREC      & 63.80         & 62.40     & \textbf{67.60}         & \textbf{66.60}     & 66.00 & 60.60         \\
    SICK-E    & \textbf{74.20}         & 73.41     & 74.06         & 75.14     & 75.34 & \textbf{75.97}         \\
    
    \bottomrule
  \end{tabular}
  \end{minipage}
\end{table*}

\begin{table*}[h]
\begin{minipage}{\linewidth}
  \caption{Sentence Evaluations on 16 benchmark datasets for Switchboard and How2 corpus. We compare the CAWE-M method with the word2vec embeddings trained with CBOW method and with CAWE-M + CBOW concatenated (Concat) embeddings.}
  \label{tab:senteval}
  \centering
  \begin{tabular}{ccccccc} %{ p{6cm}  p{1cm} p{1cm}  p{1cm} p{1cm}  }
    \toprule
    & \multicolumn{3}{c} {Switchboard} & \multicolumn{3}{c}{How2}   \\
    \cmidrule(r){2-4}\cmidrule(r){5-7}
    Dataset   & CAWE-M          & CBOW        & Concat                   & CAWE-M      & CBOW      & Concat  \\
    \midrule 
    STS 2012  & 0.3561         & \textbf{0.3639}     & 0.3470              & 0.3648     & 0.3688 & \textbf{0.3790}              \\
    STS 2013  & 0.1969         & 0.1960     & \textbf{0.2010}              & \textbf{0.2716}     & 0.2524 & 0.2675              \\
    STS 2014  & \textbf{0.3888}         & 0.3745     & 0.3795              & 0.3940     & \textbf{0.3973} & 0.3971              \\
    STS 2015  & 0.4275         & 0.4459     & \textbf{0.4481}              & 0.4173     & \textbf{0.4781} & 0.4710              \\
    STS 2016  & \textbf{0.3833}         & 0.3471     & 0.3651              & 0.3159     & \textbf{0.4023} & 0.3388              \\
    STS B     & 0.401         & \textbf{0.4100}    & 0.3995                & 0.4000     & \textbf{0.4720} & 0.4487              \\
    SICK-R    & 0.6006        & 0.6170    & \textbf{0.6228}              & 0.6440     & 0.6550 & \textbf{0.6945}               \\
    \midrule
    MR        & 64.69         & 66.24     & \textbf{66.89}               & 63.64      & 66.03 & \textbf{66.89}              \\
    MRPC      & \textbf{69.80}         & 68.99     & 68.00               & \textbf{70.61}      & 69.68 & 68.52              \\
    CR        & 72.93         & 74.49     & \textbf{75.39}               & 73.03      & \textbf{74.89} & 74.84              \\
    SUBJ      & 81.19         & \textbf{84.62}     & 84.59               & 81.01      & 84.75 & \textbf{85.04}              \\
    MPQA      & 73.75         & \textbf{76.44}     & 75.36               & 73.53      & 75.56 & \textbf{75.60}              \\
    SST       & 65.02         & 68.37     & \textbf{68.97}               & 65.13      & 67.66 & \textbf{68.20}              \\
    SST-FG    & 33.53         & 34.71     & \textbf{35.79}               & 32.08      & 33.62 & \textbf{33.67}              \\
    TREC      & 67.60         & 69.80     & \textbf{71.40}               & 60.60      & \textbf{68.40} & 67.40              \\
    SICK-E    & 74.06         & 75.02     & \textbf{76.19}               & 75.97      & 76.29 & \textbf{78.14}              \\
    \bottomrule
  \end{tabular}
  \end{minipage}
\end{table*}

%%%%%%%%%%%%%%%%%%%%%%%%%%%%%%%%%%%%%%%%
\section{Contextual Acoustic Word Embeddings}
\label{sec:cawe}
% Obtaining Acoustic Word and Sentence Embeddings
% describe method of extraction, along with peaky attention behavior

\begin{figure}
\centering
\includegraphics[width=0.5\textwidth]{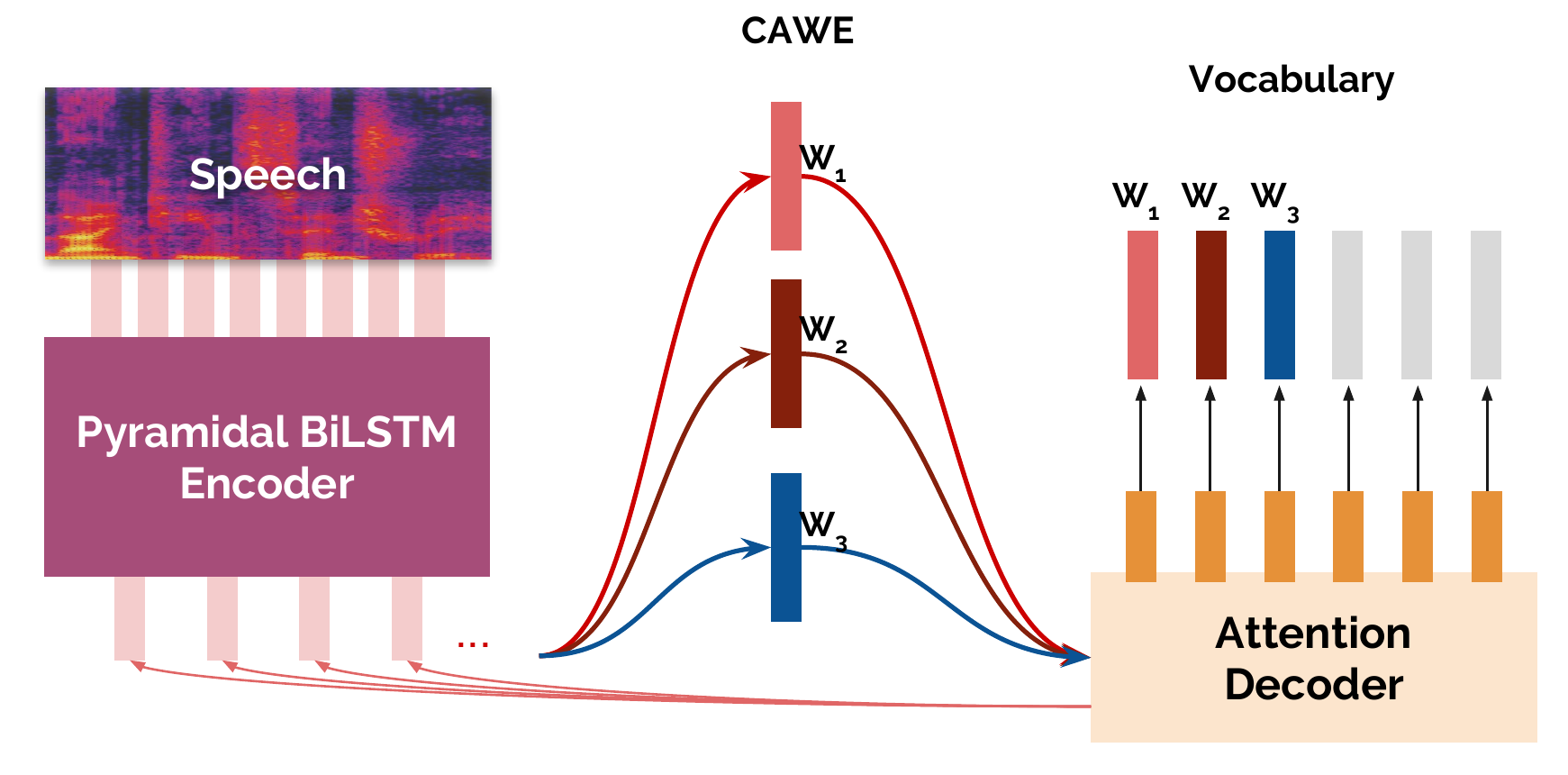}
\caption{\label{fig:cawe_model}A2W model with the CAWE representations obtained by combining the encoders representations and attention weights.}
\end{figure}

We now describe our method to obtain the acoustic word embeddings from the end-to-end trained speech recognition system described in Section \ref{sec:model_a2w}. The model is as shown in Figure \ref{fig:cawe_model} where the embeddings are constructed using the hidden representations obtained from the encoder and the attention weights from the decoder. Our method of constructing ``contextual'' acoustic word embeddings is similar to a method proposed for text embeddings, CoVe \cite{cove}. The main challenge that separates our method from CoVe \cite{cove} in learning embeddings from a supervised task, is the problem of alignment between input speech and output words. We use the location-aware attention mechanism that has the property to assign higher probability to certain frames leading to a peaky attention distribution. We exploit this property of location-aware attention in an A2W model to automatically segment continuous speech into words as shown in our previous work \cite{palaskar2018acoustic}, and then use this segmentation to obtain word embeddings. In the next two subsections, we formalize this process of constructing contextual acoustic word embeddings. Intuitively, attention weights on the acoustic frames hidden representations reflect their importance in classifying a particular word. They thereby provide a correspondence between the frame and the word within a given acoustic context. We can thus construct word representations by weighing the hidden representations of these acoustic frames in terms of their importance to the word i.e. the attention weight. We show this in the Figure \ref{fig:cawe_model} wherein the hidden representations and their attention weights are colored according to their correspondence with a particular word.

\subsection{Automatic Segmentation of Speech into Words}
Given that $a_j$ represents the acoustic frame $j$, let $encoder(a_j)$ represent the higher-level features obtained for the frame $a_j$ ( i.e. $encoder(a_j)$ = $\textbf{h} = (\textbf{h}_1,\textbf{h}_2,...,\textbf{h}_{T'})$, as explained in Section \ref{sec:model_a2w}). Then, for the $i^{th}$ word $w_i$ our model first obtains the mappings of $w_i$ to acoustic frames $a_K$ where $K$ is the set such that $ \forall k \in K$ $$k = \arg\max_j (attention(a_j))$$ over all utterances $U$ containing the word $w_i$ in the training set.

\subsection{Investigating the Utility of Attention Distribution for Constructing Embeddings: CAWE}
\label{sec:4.2baselines_cawe}
Below we describe three different ways of using attention to obtain acoustic word embeddings for a word $w_i$ (here, $n(K)$ represents the cardinality of the set $K$):
\begin{align}
     w_i &= \cfrac{\sum_{k \in K} encoder(a_k)}{n(K)}  ~\\
     w_i &= \cfrac{\sum_{k \in K} attention(a_k) \cdot encoder(a_k)}{n(K)} ~\\
     w_i &= encoder(a_k) \hspace*{0.5em}where \hspace*{0.5em} k = \arg\max_{k \in K}  attention(a_k)
\end{align}

Therefore, unweighted Average (U-AVG, Equation 1) is just the unweighted combination of all the hidden representations of acoustic frames mapped to a particular word. Attention weighted Average (CAWE-W, Equation 2) is the weighted average of the hidden representations of all acoustic frames using the attention weights for a given word. Finally, maximum attention (CAWE-M, Equation 3) is the hidden representation of the acoustic frame with the highest attention score for a given word across all utterances in the training data. We call the attention-weighted average and the maximum attention based techniques as Contextual Acoustic Word Embeddings (CAWE) since they are contextual owing to the use of attention scores (over all acoustic frames for a given word).

%%%%%%%%%%%%%%%%%%%%%%%%%%%%%%%%%%%%%%%%
\section{Experiments and Results}
\label{sec:experiments_results}
We use a commonly used speech recognition setup, the 300\,hour Switchboard corpus (LDC97S62)~\cite{switchboard_corpus} which consists of 2,430 two-sided telephonic conversations between 500 different speakers and contains 3 million words of text. Our second dataset is a 300 hour subset of the How2~\cite{how2} dataset of instructional videos, which contains planned, but free speech, often outdoor and recorded with distant microphones, as opposed to the indoor, telephony, conversational speech of Switchboard. There are 13,662 videos with a total of 3.5 million words in this corpus. The A2W obtains a word error rate of 22.2\% on Switchboard and 36.6\% on CallHome set from the Switchboard Eval2000 test set and 24.3\% on dev5 test set of How2. 

\subsection{Comparing Methods for Constructing Embeddings}
\label{ssec:baselines}
\hspace{0.5cm}\textbf{Datasets for Downstream Tasks:} We evaluate our embeddings by using them as features for 16 benchmark sentence evaluation tasks that cover Semantic Textual Similarity (STS 2012-2016 and STS B), classification: Movie Review (MR), product review (CJ), sentiment analysis (SST, SST-FG), question type (TREC), Subjectivity/Objectivity (SUBJ), and opinion polarity (MPQA), entailment and semantic relatedness using the SICK dataset for SICK-E (entailment) and SICK-R (relatedness) and paraphrase detection (MRPC). The STS and SICK-R tasks measure Spearman's coefficient of correlation between embedding based similarity and human scores, hence the scores range from $\left[-1,1\right]$ where higher number denotes high correlation. All the remaining tasks are measured on test classification accuracies. We use the SentEval toolkit~\cite{senteval} to evaluate. %\footnote{\url{https://github.com/facebookresearch/SentEval}}

\textbf{Training Details:} In all downstream evaluations involving classification tasks, we have used a simple logistic regression for classification since a better representation should lead to better scores without using complicated models (hence abstracting away model complexities from our evaluations). This also means that we can use the concatenation of CAWE and CBOW as features to the logistic regression model without adding tunable embedding parameters. 

\textbf{Discussion:} From the results in Table \ref{tab:senteval_baselines} we see that CAWE-M outperforms U-AVG by 34\% and 13\% and CAWE-W by 33.9\% and 12\% on Switchboard and How2 datasets respectively in terms of average performance on STS tasks and leads to better or slightly worse performance on the classification tasks. We observe that CAWE-W usually performs worse than CAWE-M which could be attributed to a noisy estimation of the word embeddings on the account of taking even the less confident attention scores while constructing the embedding. In contrast, CAWE-M is constructed using the most confident attention score obtained over all the occurrences of the acoustic frames corresponding to a particular word. We also observe that U-AVG performs worse than CAWE-W on STS and SICK-R tasks since it is constructed using an even noisier process in which all encoder hidden representations are weighted equally irrespective of their attention scores.

\subsection{Comparing with Text-based Embeddings}
\label{ssec:senteval}
\hspace{0.5cm}\textbf{Datasets for Downstream Tasks:} The datasets are the same as described in Section \ref{ssec:baselines}.

\textbf{Training Details:} In all the following comparisons, we compare embeddings obtained only from the training set of the speech recognition model, while the text-based word embeddings are obtained by training Continuous Bag-of-Words (CBOW) word2vec model on all the transcripts (train, validation and test). This was done to ensure a fair comparison between our supervised technique and the unsupervised word2vec method. This naturally leads to a smaller vocabulary for CAWE. Further, one of the drawbacks of A2W speech recognition model is that it fails to capture entire vocabulary, recognizing only 3044 words out of 29874 (out of which 18800 words occur less than 5 times) and 4287 out of 14242 total vocabulary for Switchboard and How2 respectively. Despite this fact, the performance of CAWE is very competitive with word2vec CBOW which does not suffer from reduced vocabulary problem. 

\textbf{Discussion:} In Table~\ref{tab:senteval}, we see that our embeddings perform as well as the text-embeddings. Evaluations using CAWE-M extracted from Switchboard based training show that the acoustic embeddings when concatenated with the text embeddings outperform the word2vec embeddings on 10 out of 16 tasks. This concatenated embedding shows that we add more information with CAWE-M that improves the CBOW embedding as well. The gains are more prominent in Switchboard as compared to the How2 dataset since How2 is planned instructional speech whereas Switchboard is spontaneous conversational speech (thereby making the How2 characteristics closer to text leading to a stronger CBOW model).

\subsection{Evaluation on Spoken Language Understanding}
\label{ssec:slu}
\hspace{0.5cm}\textbf{Dataset:} In addition to generic sentence-level evaluations, we also evaluate CAWE on the widely used ATIS dataset \cite{price1990evaluation} for Spoken Language Understanding (SLU). ATIS dataset is comprised of spoken language queries for airline reservations that have intent and named entities. Hence, it is similar in domain to Switchboard, making it a useful test bed for evaluating CAWE on a speech-based downstream evaluation task. 

\textbf{Training Details:} For this task, our model is similar to the simple Recurrent Neural Network (RNN) based model architecture as investigated in \cite{bengio_slu}. Our architecture is comprised of an embedding layer, a single layer RNN-variant (Simple RNN, Gated Recurrent Unit (GRU)) along with a dense layer and softmax. In each instance, we train our model for 10 epochs with RMSProp (learning rate 0.001). We train each model 3 times with different seed values and report average performance.

\begin{table}[h]
  \caption{Speech-based contextual word embeddings (CAWE-M and CAWE-W) match the performance of the text-based embeddings (CBOW) on the ATIS dataset with an RNN and GRU model}
  \label{tab:slu}
  \centering
  \begin{tabular}{cccc}
    \toprule
            & CAWE-M    & CAWE-W    & CBOW    \\
    \midrule    
    RNN     & 91.49     & 91.67     & 91.82      \\
    GRU     & 93.25     & 93.56     & 93.11 \\
    \bottomrule
  \end{tabular}
\end{table}

\textbf{Discussion:} \cite{bengio_slu} concluded that text-based word embeddings trained on large text corpora consistently lead to better performance on the ATIS dataset. We demonstrate that direct speech-based word embeddings could lead to matching performance when compared to text-based word embeddings in this speech-based downstream task, thus highlighting the utility of our speech based embeddings. Specifically, we compare the test scores obtained by initializing the model with CAWE-M, CAWE-W and CBOW embeddings and fine-tuning them based on the task.

\section{Conclusion}
\label{sec:conclusion}
We present a method to learn contextual acoustic word embeddings from a sequence-to-sequence acoustic-to-word speech recognition model that learns to jointly segment and classify speech. We analyze the role of attention in constructing contextual acoustic word embeddings, and find our acoustic embeddings to be highly competitive with word2vec (CBOW) text embeddings. We discuss two variants of such contextual acoustic word embeddings which outperform the simple unweighted average method by upto 34\% on semantic textual similarity tasks. The embeddings also matched the performance of text-based embeddings in spoken language understanding, showing the use of this model as a pre-trained model for other speech-based downstream tasks. We surmise that contextual audio embeddings will generalize and improve downstream tasks in a way that is similar to their text counterparts, despite the additional complexity presented by noisy audio input. In the future, we will explore ways to scale our model to larger corpora, larger vocabularies and compare with non-contextual acoustic word embedding methods. %\footnote{We have submitted a longer version of this paper to the NIPS IRASL 2018 non-archival workshop. If accepted at ICASSP we will retract the paper from IRASL and adhere to IEEE copyright requirements.}

% Florian - is this past tense or present tense? I changed it all to present tense. We might also want to repeat our most important numeric results here - i.e. "our word embeddings improve X by Y%", just to drive home this message
%into individual words. Using this property of automatic segmentation, we present different methods to learn acoustic word embeddings from a speech recognition model.
%This is particularly challenging for speech since unlike machine translation, which only has to learn alignments between words, we have to additionally learn alignments between multiple acoustic frames and words. Despite the additional challenges, our demonstration of utility word embeddings obtained directly from a supervised speech recognition model can serve as a precursor to speech recognition models being extensively used for downstream tasks like CoVe in natural language processing and VGG in computer vision. 

\section{Acknowledgements}
\label{sec:acknowledgements}
This work was supported by the Center for Machine Learning and Health (CMLH) at Carnegie Mellon University and by Facebook.

\bibliographystyle{IEEEbib}
\bibliography{strings,refs}

\end{document}